\documentclass[a4paper, 10pt, conference]{ieeeconf}      
\usepackage{FG2021}
\usepackage{amsmath,amssymb,amsfonts}
\usepackage{array}
\usepackage{booktabs}
\usepackage{cite}
\usepackage{graphicx}
\usepackage{multirow}
\usepackage{subcaption}
\usepackage{times}
\usepackage{epsfig}
\usepackage{graphicx}
\usepackage{multirow}
\usepackage{epsfig}
\usepackage[accsupp]{axessibility} 

\newcolumntype{L}{>{$}l<{$}}
\newcolumntype{C}{>{$}c<{$}}
\newcolumntype{R}{>{$}r<{$}}

\newcommand{\nm}[1]{\textnormal{#1}}
\newcommand{\etal}{\emph{et~al.}}
\newcommand{\norm}[1]{\left\lVert #1 \right\rVert}

\FGfinalcopy 

\IEEEoverridecommandlockouts                              
\def\FGPaperID{363} 

\title{\LARGE \bf
Synthesis-Guided Feature Learning for \\Cross-Spectral Periocular Recognition
}

\author{\parbox{16cm}{\centering
    {\large Domenick Poster and Nasser Nasrabadi}\\
    {\normalsize
    Lane Dept. of Computer Science and Electrical Engineering, West Virginia University, Morgantown, USA\\}}
}

\begin{document}

\ifFGfinal
\thispagestyle{empty}
\pagestyle{plain}
\else
\author{Anonymous FG2021 submission\\ Paper ID \FGPaperID \\}
\pagestyle{plain}
\fi
\maketitle

\begin{abstract}
A common yet challenging scenario in periocular biometrics is cross-spectral matching - in particular, the matching of visible wavelength against near-infrared (NIR) periocular images. We propose a novel approach to cross-spectral periocular verification that primarily focuses on learning a mapping from visible and NIR periocular images to a shared latent representational subspace, and supports this effort by simultaneously learning intra-spectral image reconstruction. We show the auxiliary image reconstruction task (and in particular the reconstruction of high-level, semantic features) results in learning a more discriminative, domain-invariant subspace compared to the baseline while incurring no additional computational or memory costs at test-time. The proposed Coupled Conditional Generative Adversarial Network (CoGAN) architecture uses paired generator networks (one operating on visible images and the other on NIR) composed of U-Nets with ResNet-18 encoders trained for feature learning via contrastive loss and for intra-spectral image reconstruction with adversarial, pixel-based, and perceptual reconstruction losses. Moreover, the proposed CoGAN model beats the current state-of-art (SotA) in cross-spectral periocular recognition. On the Hong Kong PolyU benchmark dataset, we achieve 98.65\% AUC and 5.14\% EER compared to the SotA EER of 8.02\%. On the Cross-Eyed dataset, we achieve 99.31\% AUC and 3.99\% EER versus SotA EER of 4.39\%.

\end{abstract}

\section{INTRODUCTION}
\begin{figure}[t]
    \centering
      \begin{subfigure}{\columnwidth}
      \centering
        \includegraphics[height=40pt]{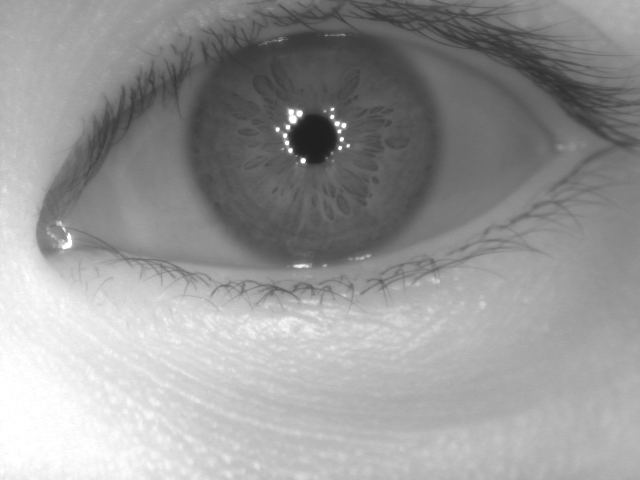}
        \includegraphics[height=40pt]{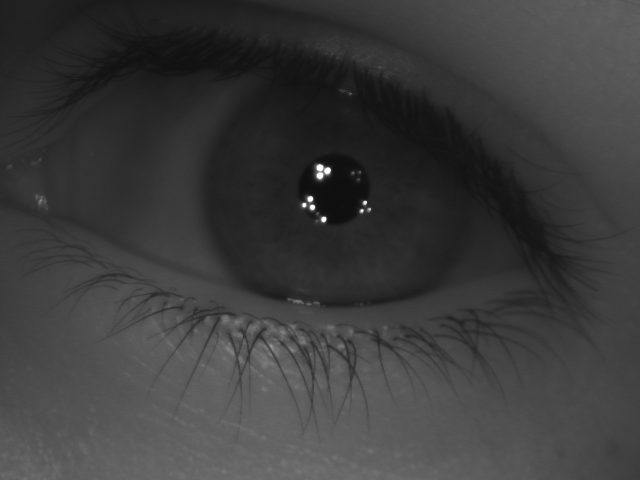}
        \includegraphics[height=40pt]{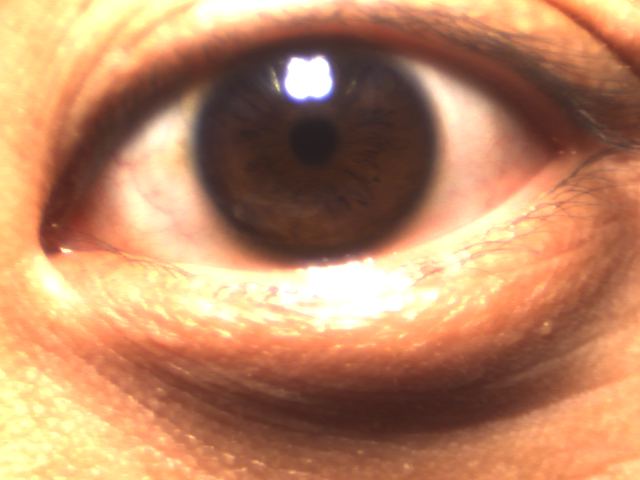}
        \includegraphics[height=40pt]{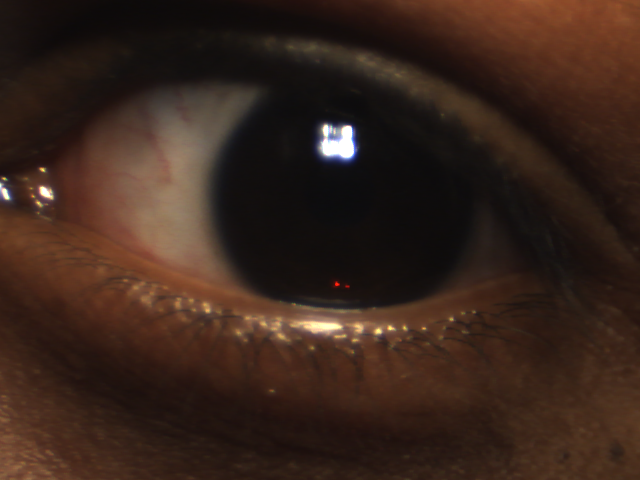}
      \end{subfigure}
      \par\medskip
      \begin{subfigure}{\columnwidth}
      \centering
        \includegraphics[height=40pt]{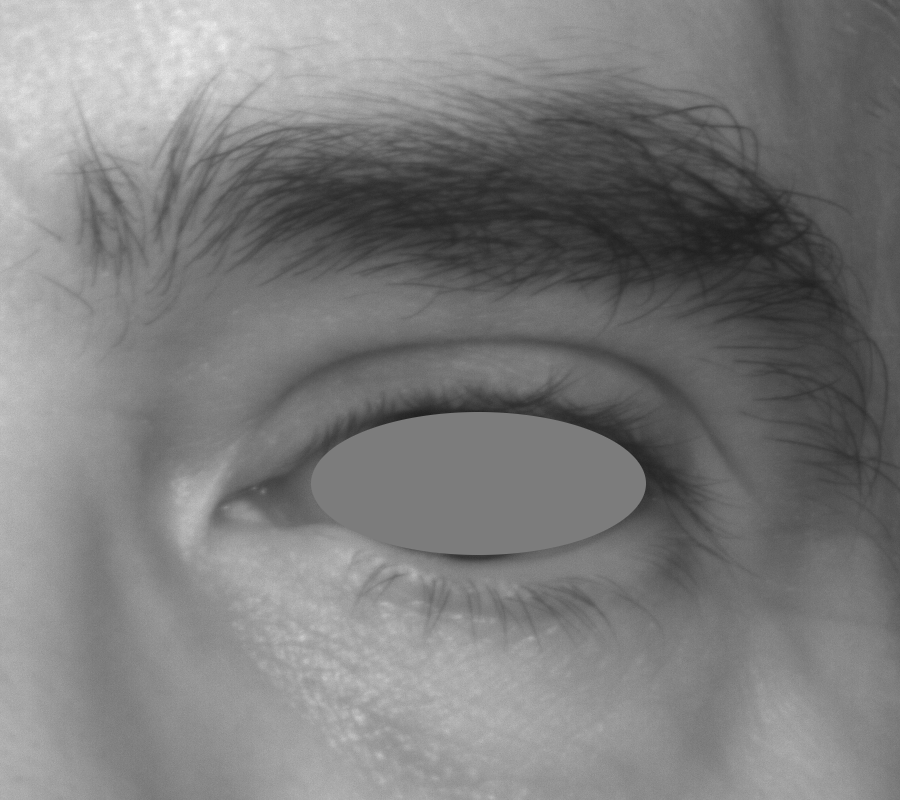}
        \includegraphics[height=40pt]{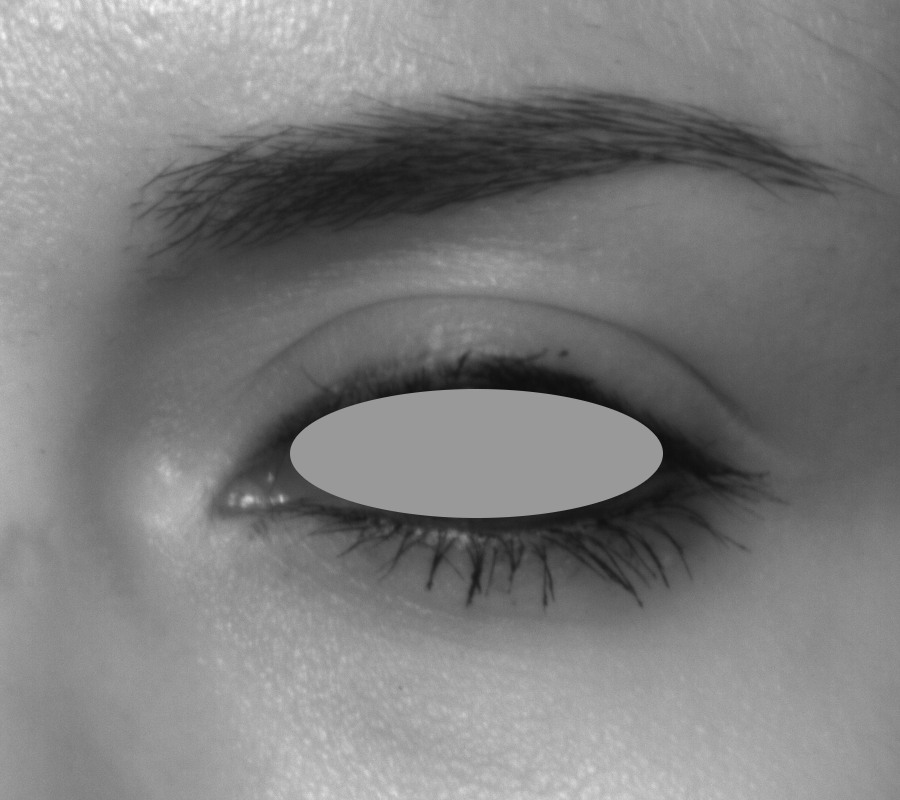}
        \includegraphics[height=40pt]{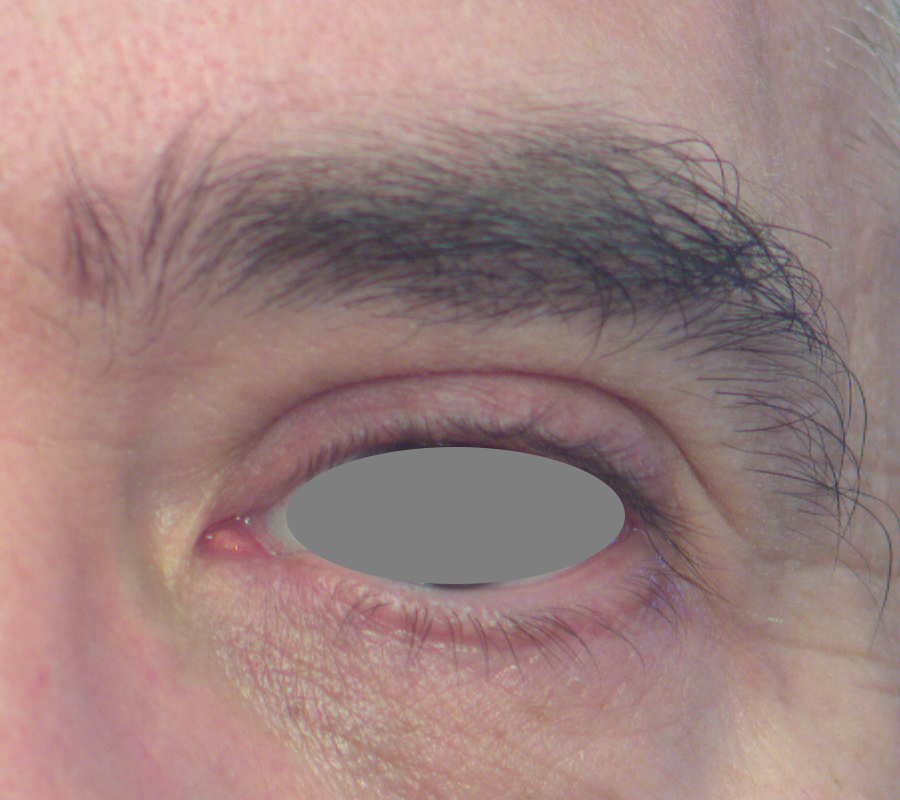}
        \includegraphics[height=40pt]{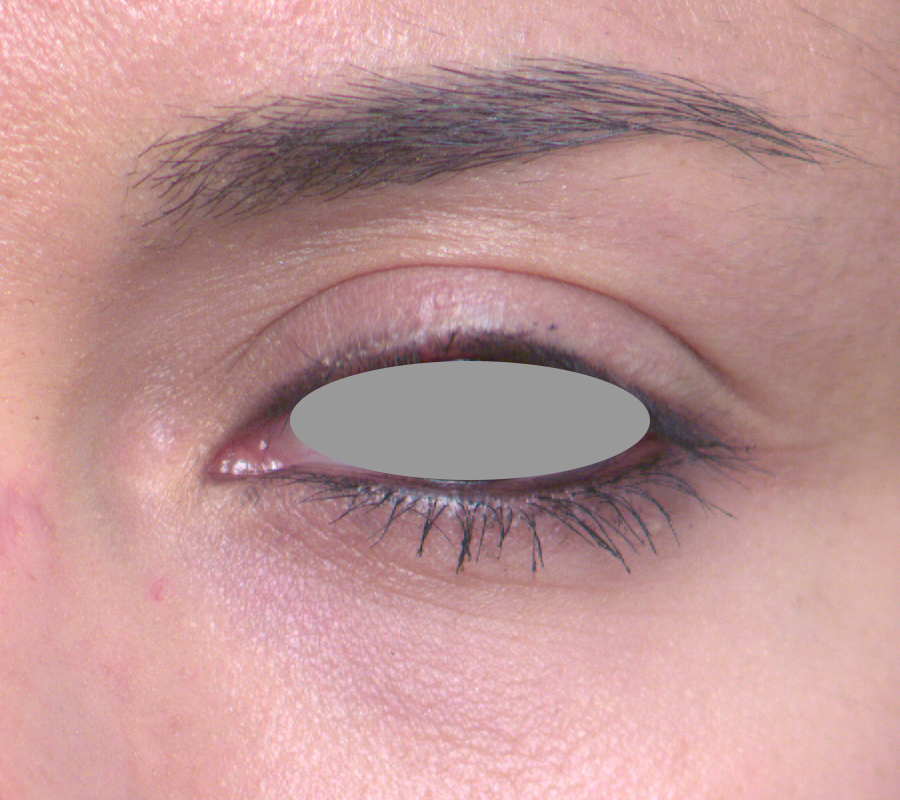}
      \end{subfigure}
\caption{Near-infrared (left) and visible (right) periocular images from HK PolyU (top) and Cross-eyed (bottom) datasets.}
\label{fig:small_samples}
\end{figure}

Periocular recognition uses the region surrounding the eye to recognize individuals. The periocular region can offer a good trade-off in terms of recognition accuracy versus useability. Cross-spectral recognition further relaxes the constraints imposed on input images at the cost of introducing a challenging domain gap between images of different spectral wavelengths. However, this proposition is made more attractive for periocular recognition due to the prevalence of iris recognition systems which typically generate near-infrared (NIR) periocular images, and also to the increased use of masks being worn which partially obscure the face. Examples of periocular images are shown in Fig. \ref{fig:small_samples}.

Two common strategies to address the cross-spectral domain-shift are to 1) map images of either domain to a shared representational subspace \cite{chopra2005learning,hernandez2019cross,zanlorensi2019deep,behera2020twin}, or 2) translate images from one domain to the other \cite{hernandez2020cross}, typically to use an existing uni-modal recognition model. Drawing on elements of both strategies, we propose a novel approach to cross-spectral periocular recognition which utilizes coupled convolutional neural networks (CNN) to perform shared latent subspace learning and feature matching while simultaneously leveraging Conditional Generative Adversarial Networks \cite{goodfellow2014generative,mirza2014conditional} (cGAN) to guide the learning process. Our Coupled GAN (CoGAN) architecture outperforms the state-of-the-art on the Hong Kong PolyU \cite{nalla2016toward} and Cross-Eyed \cite{sequeira2016cross} cross-spectral periocular recognition datasets.

The primary contributions of this work are:
\begin{itemize}
    \item A novel cross-spectral periocular recognition approach utilizing the proposed CoGAN architecture.
    \item A set of benchmark experiments on the Hong Kong PolyU and Cross-Eyed cross-spectral periocular datasets validating the performance of the proposed approach over baseline frameworks and state-of-the-art.
    \item A series of ablation studies validating the benefits of the auxiliary goal of image reconstruction to shared subspace feature learning.
\end{itemize}

\section{RELATED WORKS}
\begin{figure*}[t]
    \centering
    \includegraphics[width=0.72\linewidth]{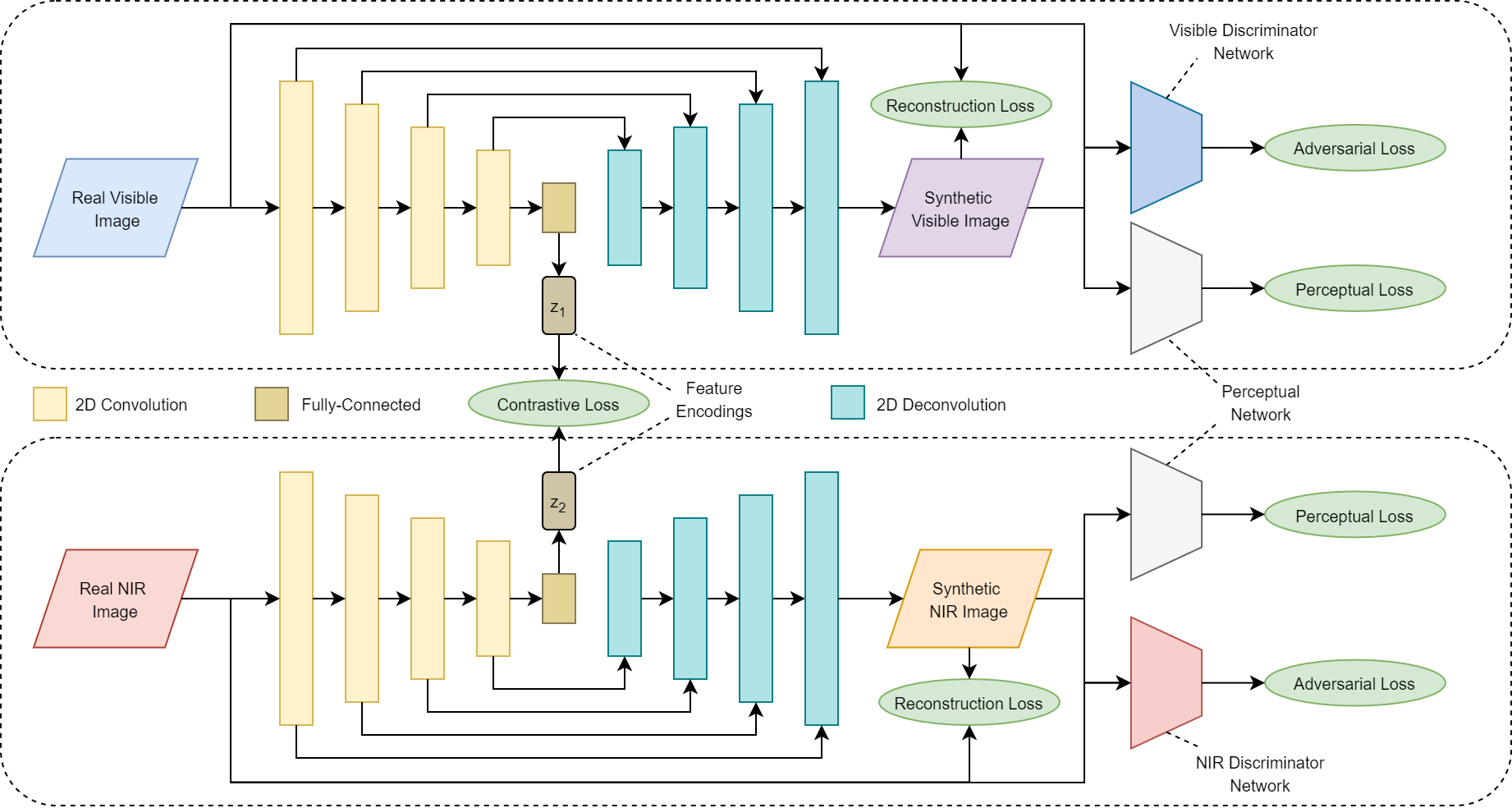}
    \caption{The CoGAN architecture composed of coupled ResNet-18 encoders (yellow) embedded within dual U-Net cGANs.}
    \label{fig:arch}
\end{figure*}

Deep-learning based approaches have been used in a variety of periocular biometric scenarios including intra-spectral recognition \cite{hernandez2019cross,zanlorensi2019deep,hernandez2018periocular,kumari2020periocular}, bi-modal fusion \cite{zanlorensi2019deep}, and cross-spectral recognition \cite{zanlorensi2019deep,behera2020twin,hernandez2020cross}. In cross-spectral periocular recognition, near perfect or perfect results have been attained \cite{zanlorensi2019deep,behera2020twin,hernandez2020cross} on the Hong Kong PolyU \cite{nalla2016toward} and Cross-Eyed \cite{sequeira2016cross} \textit{closed world} protocols due to the high correlation between the \textit{non-class-disjoint} train and test data. However, Zanlorensi \etal \cite{zanlorensi2019deep} also achieve state-of-the-art performance on the more challenging \textit{class-disjoint} \textit{open world} protocol using coupled ResNet-50 \cite{he2016deep} networks trained for feature extraction and matching. Our approach primarily differs in that we enhance the subspace feature learning by simultaneously training for image reconstruction in a multi-task fashion while also using a much smaller ResNet-18 \cite{he2016deep} network.

A generative adversarial network (GAN) \cite{goodfellow2014generative} is composed of a generator network $G(z)$ which generates data given a random input vector $z$, and a discriminator network $D(\cdot)$ which outputs the probability that the input came from the training data $x$ or from the generator. These two networks compete by playing a ``minimax game." In a cGAN \cite{mirza2014conditional}, the networks are conditioned on some auxiliary information $y$. The generator function takes the form $G(z|y)$ instead of $G(z)$. The adversarial loss for the cGAN is:
\begin{multline}\label{eq:adv}
\min_G \max_D \mathcal{L}_{cGAN}(G;D;x;y) = E_{x\sim P_{data}(x)}[log(D(x|y))] \\ + E_{z\sim P_{z}(z)}[log(1-D(G(z|y)))].
\end{multline}

Several methods have been proposed to refine the quality of the generated images. U-Nets \cite{ronneberger2015u} are generators with skip connections that forward lower-level features from the encoder layers to their mirrored counterparts in the decoder. Pix2Pix \cite{isola2017image} performs image-to-image translation and style transfer using a U-Net cGAN with an L1 reconstruction constraint between synthesized and target images and a patch-based discriminator (PatchGAN) designed to model the local, high-frequency nuances of the data. While Pix2Pix models local and fine-grained image statistics, perceptual loss \cite{johnson2016perceptual} has been proposed to measure the difference in the high-level, semantic features of images through the use of a pre-trained, fixed-loss ``perceptual network" such as VGG16 \cite{simonyan2014very} pretrained on ImageNet \cite{deng2009imagenet}. Different from Pix2Pix, our approach uses a combination of global L2 reconstruction loss and perceptual loss to emphasize the high-level semantic features more likely to be present in both domains over the reconstruction of domain-specific local features.

Converting near-infrared periocular images to visible wavelength images or vice-versa enables the use of intra-spectral recognition algorithms. Hernandez-Diaz \etal \cite{hernandez2020cross} employ Pix2Pix for this purpose, showing its effectiveness for data augmentation, but ultimately achieve lower performance than a direct feature-learning approach \cite{zanlorensi2019deep}. Alternatively, Taherkhani \etal \cite{taherkhani2020pf} uses coupled cGANs for extreme off-pose face recognition, demonstrating its potential for learning multi-modal representations.

\label{sec:rel}
\section{COUPLED GAN}

The overall architecture of the proposed Coupled GAN (CoGAN) is primarily composed of two cGANs. One cGAN processes visible-wavelength images while the other processes NIR images. The networks are ``coupled" by the joint task of learning a mapping from periocular images to feature representations in a common latent subspace. While our ultimate goal is to learn discriminative and domain-invariant feature encodings for cross-spectral periocular recognition, we propose to simultaneously optimize for a set of auxiliary objectives related to image reconstruction. We hypothesize these secondary reconstruction tasks can help the feature learning process key into important visual features.

Fig. \ref{fig:arch} illustrates the CoGAN's main components. The generators have a U-Net \cite{ronneberger2015u} encoder-decoder structure with ResNet-18 \cite{he2016deep} encoders (minus the final softmax layer) followed by deconvolutional decoder layers. The discriminators are VGG-like \cite{simonyan2014very} 4-layer CNNs (with filters sizes of 16, 32, 64, and 128, respectively) followed by a fully-connected layer with a single scalar output. The perceptual network is an ImageNet pre-trained VGG16 network. Once the model has been trained, the encoders can operate in isolation from the rest of the CoGAN.
\subsubsection{Shared Feature Subspace Learning}
The ultimate goal of the Coupled GAN is to conduct cross-spectral matching by learning discriminative, domain-invariant features. The contrastive loss \cite{chopra2005learning} term is the portion of the objective function which is most directly related to our ultimate goal. All other losses are auxiliary to the learning process. 

Minimizing the contrastive loss encourages the features extracted from a pair of genuine images (i.e. instances of the same class) to be similar and the features of imposter pairs (instances of different classes) to be least some margin $m$ apart. Let $X_V=\{x_V^i\}_{i=1}^{N}$ and $X_I=\{x_I^j\}_{j=1}^{N}$ be the visible wavelength and near-infrared periocular images, respectively. Let $\mathcal{D}_z(x^{i}_{V};x^{j}_{I})$ be the distance measure between the features extracted from a pair of images. Given the feature encodings $z_1(x_V^i)$ and $z_2(x_I^j)$ obtained from the encoder sub-networks, we calculate the distance measure as the L2-norm:
\begin{equation}
\mathcal{D}_z(x^{i}_{V};x^{j}_{I}) = \norm{z_1(x^{i}_{V}) - z_2(x^{j}_{I})}_{2}.
\end{equation}
Let us define ground-truth labels $y^{ij}=0$ for genuine pairs and $y^{ij}=1$ for imposter pairs. The contrastive loss function is then written as:
\begin{multline}
\mathcal{L}_C(x^{i}_{V};x^{j}_{I};y^{ij}) = (1 - y^{ij})\frac{1}{2}(\mathcal{D}_z(x^i_V,x^j_I)^2 \\ + (y^{ij})\frac{1}{2}(\text{max}(0, m - \mathcal{D}_z(x^i_V,x^j_I)))^2,
\end{multline}
\begin{equation}
\mathcal{L}_C^* = \frac{1}{N^2}\sum_{i=1}^{N}\sum_{j=1}^{N}\mathcal{L}_C(x^{i}_{V}, x^{j}_{I}, y^{ij}),
\end{equation}
where $\mathcal{L}^*$ denotes a combined loss computed over both visible and NIR domains.
\subsection{Image Synthesis}
Instead of performing cross-spectral image synthesis, we utilize the pair of cGANs to refine the feature learning process via three auxiliary loss terms: adversarial loss, reconstruction loss, and perceptual loss; all of which work in concert with the contrastive loss to help guide the encoders in learning important visual features.
\subsubsection{Adversarial Loss}
Using (\ref{eq:adv}) for the adversarial loss of a single cGAN, we condition on the input images in order to recreate them. Let $G_V$ and $G_I$ be generators operating on visible and IR images, respectively, while $D_V$ and $D_I$ are their respective discriminators. The total adversarial loss for the coupled networks is:
\begin{multline}
\mathcal{L}_A^* = \frac{1}{N^2}((\sum_{i=1}^{N}\mathcal{L}_{cGAN}(G_V,D_V,x_V^i,x_V^i)) \\ + (\sum_{j=1}^{N}\mathcal{L}_{cGAN}(G_I,D_I,x_I^j,x_I^j))).
\end{multline}
\subsubsection{Reconstruction Loss}
Minimizing the reconstruction loss trains the GAN to recreate low-level, fine-grained features. We use the L2 distance between the pixel values of a given image $x$ and its synthesized version $G(x)$ as the reconstruction loss. The total reconstruction loss is defined as follows:
\begin{equation}
    \mathcal{L}_R(x;G) = \norm{x - G(x)}_2^2,
\end{equation}
\begin{equation}
\mathcal{L}_R^* = \frac{1}{N^2}(\sum_{i=1}^{N}\mathcal{L}_R(x^i_{V},G_{V}) + ( \sum_{j=1}^{N}\mathcal{L}_R(x^j_{I},G_{I}))).
\end{equation}
\subsubsection{Perceptual Loss}
Given an image $x$ and its reconstruction, the perceptual loss \cite{johnson2016perceptual} measures the distance between the high-level features extracted by the $k$th layer of a fixed loss network $\phi$. Let $\phi_k(\cdot)$ be the $C_k\times W_k \times H_k$ activations of the $k$th network layer for a given input image. The perceptual loss is the L2 distance between the feature representations:
\begin{equation}\label{eq:perc}
    \mathcal{L}_P(x;G) = \frac{1}{CWH}\sum_{c=1}^{C_k}\sum_{w=1}^{W_k}\sum_{h=1}^{H_k}\norm{\phi_k(x) - \phi_k(G(x))}_2^2,
\end{equation}
\begin{equation}
\mathcal{L}^*_P = \frac{1}{N^2}(\sum_{i=1}^{N}\mathcal{L}_P(x^i_{V},G_V) + \sum_{j=1}^{N}\mathcal{L}_P(x^j_{I},G_I)).
\end{equation}
 We use the activations of the last convolutional layer of a pre-trained VGG16 network for $\phi_k$.
\subsection{Training and Implementation}\label{sec:implementation}
The overall objective function of the Coupled GAN is the summation of the multi-task losses plus an additional L2 weight decay term:
\begin{equation}
\mathcal{L} = \lambda_C\mathcal{L}^*_C + \lambda_A\mathcal{L}^*_A + \lambda_R\mathcal{L}^*_R + \lambda_P\mathcal{L}^*_{P} + \lambda_{L2}\mathcal{L}^*_{L2},
\end{equation}
where $\lambda$ represents the coefficients of the individual loss terms. This function is minimized by Stochastic Gradient Descent via the Adam optimizer.

We set $\lambda_A=\lambda_R=\lambda_P=1.0$. Optimal values for $\lambda_C$ vary slightly depending on the dataset (see Section \ref{sec:ablations}). The learning rate and L2 weight decay ($\lambda_{L2}$) are fixed at $1\times10^{-4}$. Training is done for 300 epochs, with every mini-batch composed of 100 NIR-visible image pairs, resized to $256\times256$, for a total mini-batch size of 200 images. Each pair has a 50\% chance of being either a genuine or imposter pair. A 5-fold cross validation scheme on the training data was employed to tune hyperparameters and determine the number of training epochs. The models were implemented with Pytorch 1.7.0 on machines with 2x Tesla V100 GPUs. The code has been made available at https://github.com/vonclites/cogan.
\section{DATASETS}
\begin{table*}[t]
\centering
\caption{Open world protocol of the Hong Kong PolyU and Cross-Eyed cross-spectral periocular datasets.}
\label{tab:datasets}
\begin{tabular}{LLLLL}
\toprule
\multicolumn{1}{l}{Dataset} &
\multicolumn{1}{c}{Train/Test Subjects} &
\multicolumn{1}{c}{Train/Test Classes} &
\multicolumn{1}{c}{Train/Test Images} &
\multicolumn{1}{c}{Gen/Imposter Test Pairs} \\

\midrule

\nm{HK PolyU} & 104.5/104.5 & 209/209 & 6,270/6,270 & 21,945/9,781,200 \\
\nm{Cross-Eyed} & 60/60 & 120/120 & 1,920/1,920 & 3,360/913,920 \\
\bottomrule
\end{tabular}
\end{table*}

Our approach is benchmarked on the \textit{open world} protocols of Hong Kong PolyU \cite{nalla2016toward} and Cross-Eyed \cite{sequeira2016cross} datasets following the method used in \cite{zanlorensi2019deep}. Dataset details are provided in Table \ref{tab:datasets} and example images in Fig. \ref{fig:small_samples}. Following  standard practice in periocular recognition, we consider a subject’s left and right eyes to be unique classes. Verification is conducted using a one-against-all pairwise matching strategy.

The Hong Kong PolyU \cite{nalla2016toward} dataset is composed of simultaneously acquired images in the NIR and visible spectrums. The entire database has 12,540 images with a resolution of 640×480. In both visible and thermal spectrums, there are 15 samples of each eye (left and right) from 209 subjects (418 classes). The first 104 subjects plus the left eye of the 105th are assigned to the training set and the rest to testing.

The Cross-Eyed \cite{sequeira2016cross} dataset has 3,840 images from 120 subjects (240 classes). There are eight samples from each of the classes for each spectrum. All images are 400×300 resolution and were obtained at a distance of 1.5 meters. The first 60 subjects are assigned to the training set and the remaining 60 to the test set.

Images are rescaled to $256\times256$ with zero mean and unit variance per training data statistics. For data augmentation, training images are zero-padded to $272\times272$ and a $256\times256$ crop is extracted.

\section{RESULTS AND DISCUSSION}
In this section, we present and discuss the benchmark results of the CoGAN framework along with additional experiments and ablation studies investigating the benefits of our hybrid approach. Once trained, the only components which are utilized for evaluating the CoGAN's performance are the encoder sub-networks. Performance is evaluated in terms of the Area Under the ROC Curve (AUC) and Equal Error Rate (EER). Also provided is the False Rejection Rate at False Accept Rates of 1\% and 10\% (FRR@FAR=1\%, FRR@FAR=10\%).

\subsection{Comparison with State of the Art}
At the time of writing, the only recent work which reports results on the \textit{open world} protocols of the benchmark datasets is, to the best of our knowledge, Zanlorensi \etal \cite{zanlorensi2019deep}, who have achieved state-of-the-art performance on the cross-spectral scenario. In Table \ref{tab:comp}, the performance of the proposed CoGAN model is compared with the state-of-the-art and two baselines models. The baseline versions are equivalent to training the CoGAN using only contrastive loss and weight decay. The CoGAN achieves improved EER(\%) over the ResNet18 baseline and the larger, coupled ResNet-50 model of \cite{zanlorensi2019deep}, which in turn performs better than our ResNet-50 baseline, potentially due to their use of cosine similarity. ResNet-18 can also be seen to outperform ResNet-50, consistent with the intra-spectral periocular recognition results reported in \cite{kumari2020periocular}.
\begin{table}[t]
\centering
\caption{Comparison with State-of-the-Art and baseline.}
\label{tab:comp}
\begin{tabular}{LLLLLL}
\toprule
\multicolumn{1}{l}{Dataset} &
\multicolumn{1}{l}{Model} &
\multicolumn{1}{c}{AUC\%} &
\multicolumn{1}{c}{EER\%} &
\multicolumn{2}{c}{FRR@FAR} \\

\cmidrule(lr){5-6}
\multicolumn{0}{c}{}   &
\multicolumn{0}{c}{}   &
\multicolumn{0}{c}{}   &
\multicolumn{0}{c}{}   &
\multicolumn{1}{c}{1\%} &
\multicolumn{1}{c}{10\%} \\
\midrule

{\multirow{4}{*}{HK PolyU}} &
\nm{ResNet-50} & 96.03 & 9.93 & 31.22 & 11.25 \\ &
\nm{Zanlorensi \cite{zanlorensi2019deep}} & \multicolumn{1}{c}{-} & 8.02 & \multicolumn{1}{c}{-} & \multicolumn{1}{c}{-} \\ &
\nm{ResNet-18} & 98.16 & 5.85 & 18.6 & 4.36 \\ &
\nm{\textbf{CoGAN}} & \mathbf{98.65} & \mathbf{5.14} & \mathbf{12.32} & \mathbf{3.27} \\
\cmidrule(lr){1-6}
{\multirow{4}{*}{Cross-Eyed}} &
\nm{ResNet-50} & 95.43 & 10.00 & 26.43 & 10.04 \\ &
\nm{Zanlorensi \cite{zanlorensi2019deep}} & \multicolumn{1}{c}{-} & 4.39 & \multicolumn{1}{c}{-} & \multicolumn{1}{c}{-} \\ &
\nm{ResNet-18} & 98.95 & 4.15 & 10.19 & 2.62 \\ & 
\nm{\textbf{CoGAN}} & \mathbf{99.41} & \mathbf{3.07} & \mathbf{6.11} & \mathbf{1.66} \\
\bottomrule
\end{tabular}
\end{table}


\subsection{Ablation Studies}\label{sec:ablations}
The following ablation studies serve to further evaluate the performance of the CoGAN and the effect of the auxiliary image synthesis tasks on the end-goal of periocular verification. Table \ref{tab:abl} lists the AUC(\%) for various sizes of the feature encoding vectors ($z_1$ and $z_2$) and contrastive cost coefficients ($\lambda_C$). Similar to \cite{zanlorensi2019deep}, we also found the optimal feature size to be smaller for the Hong Kong dataset than the Cross-Eyed.

\begin{table}[t]
\centering
\caption{Feature Dimension Size $|z|$ vs. Contrastive Cost Coefficient $\lambda_C$ in terms of AUC(\%). $\lambda_A$, $\lambda_R$, $\lambda_P$ set to 1.0.}
\label{tab:abl}
\begin{tabular}{LLLLLL}
\toprule
\multicolumn{1}{l}{Dataset} &
\multicolumn{1}{l}{$|z|$} &
\multicolumn{4}{c}{$\lambda_C$} \\

\cmidrule(lr){3-6}

\multicolumn{0}{c}{}   &
\multicolumn{0}{c}{}   &
\multicolumn{1}{c}{1.0}   &
\multicolumn{1}{c}{2.0}   &
\multicolumn{1}{c}{\textbf{5.0}}   &
\multicolumn{1}{c}{10.0}   \\
\midrule

{\multirow{1}{*}{HK PolyU}} & \nm{32} & 97.91 & 97.85 & 98.16 & 97.89 \\
& \nm{\textbf{64}} & 98.35 & 98.21 & \mathbf{98.65} & 98.42 \\
& \nm{128} & 98.25 & 97.95 & 97.87 & 97.92 \\
& \nm{256} & 98.28 & 98.54 & 98.09 & 98.38 \\
\cmidrule(lr){1-6}
{\multirow{1}{*}{Cross-Eyed}} & 32 & 98.69 & 98.74 & 98.94 & 98.88 \\
& 64 & 98.94 & 98.84 & 98.58 & 98.94 \\
& \textbf{128} & 99.01 & 99.24 & 99.31 & \mathbf{99.41} \\
& 256 & 98.08 & 97.63 & 97.88 & 98.03 \\
\bottomrule
\end{tabular}
\end{table}

\begin{table}[t]
\centering
\caption{The effect of the auxiliary image synthesis tasks (adversarial loss $\lambda_A$, reconstruction loss $\lambda_R$, and perceptual loss $\lambda_P$) on the HK PolyU dataset.}
\label{tab:synth_abl}
\begin{tabular}{LLLLL}
\toprule
\multicolumn{1}{c}{$\lambda_C$} &
\multicolumn{1}{c}{$\lambda_A$} &
\multicolumn{1}{c}{$\lambda_R$} &
\multicolumn{1}{c}{$\lambda_P$} &
\multicolumn{1}{c}{AUC, \%} \\

\midrule
1.0 & 0.0 & 0.0 & 0.0 & 98.41 \\
5.0 & 1.0 & 0.0 & 0.0 & 98.54 \\
5.0 & 1.0 & 1.0 & 0.0 & 98.55 \\
5.0 & 1.0 & 0.0 & 1.0 & 98.63 \\
5.0 & 1.0 & 1.0 & 1.0 & \mathbf{98.65} \\
\bottomrule
\end{tabular}
\end{table}

Table \ref{tab:synth_abl} shows the gradual improvements gained by incorporating the additional loss terms of the GAN. The first row of the table represents a baseline version of the model constrained only by the primary contrastive loss objective. We can see that the largest impact over the baseline is made by a combination of adversarial and perceptual loss. This suggests that by emphasizing the importance of capturing the high-level, global appearance, the CoGAN is better able to guide the learning of discriminative features.

\section{CONCLUSION}
We show our novel approach to cross-spectral periocular recognition achieves state-of-the-art results using the proposed CoGAN architecture. Our experiments demonstrate that by introducing auxiliary intra-spectral image reconstruction tasks to support the effort of shared subspace feature learning for cross-spectral periocular recognition, the CoGAN attains higher performance over a baseline version of the model. Our work presents the CoGAN architecture as a promising framework for further research in both intra-spectral and cross-spectral periocular recognition, given the compact nature of its ResNet-18 backbone, lack of need for well-aligned image pairs, and potential for cross-spectral synthesis applications. In particular, our investigation reveals that perceptual loss may aid in disentangling domain-specific information from the feature representations used for matching, indicating a potential direction for future work.



{\small
\bibliographystyle{ieee}
\bibliography{refs}
}

\end{document}